\documentclass{article}

\usepackage[T1]{fontenc}
\usepackage{CJKutf8} 
\usepackage{iclr2027_conference,times}

\iclrfinalcopy

\usepackage{graphicx}
\usepackage[section]{placeins}
\usepackage{booktabs}
\usepackage{amsmath}
\usepackage{amssymb}
\usepackage{xcolor}
\usepackage[font=small,labelfont=bf]{caption}
\usepackage{microtype}
\usepackage{algorithm}
\usepackage{algpseudocode}
\usepackage{enumitem}

\newif\ifauthornotes
\authornotesfalse

\usepackage{multirow}
\usepackage[colorlinks=true,linkcolor=blue!45!black,citecolor=blue!45!black,
            urlcolor=blue!45!black]{hyperref}

\definecolor{strongc}{HTML}{C1121F}
\definecolor{weakc}{HTML}{0077B6}

\algrenewcommand\algorithmicindent{1.0em}

\hypersetup{pdftitle={ALLSPARK: WEAK TO STRONG TRANSFER
VIA ALTERNATING CHAINS OF THOUGHT},
            pdfauthor={Kaizhao Liang, Junxiong Wang, Chen Liang, Zhendong Wang, Qiang Liu}}

\title{Allspark: Weak to Strong Transfer\\ via Alternating Chain of Thought}
\author{Kaizhao Liang\\
UT Austin\\
{\footnotesize\texttt{kaizhaol@utexas.edu}}
\And
Junxiong Wang\\
Together AI\\
{\footnotesize\texttt{junxiong@cs.cornell.edu}}
\And
Chen Liang\\
Microsoft\\
{\footnotesize\texttt{chenliang1@microsoft.com}}
\AND
Zhendong Wang\\
Microsoft\\
{\footnotesize\texttt{zhendong.wang@utexas.edu}}
\And
Qiang Liu\\
UT Austin\\
{\footnotesize\texttt{lqiang@cs.utexas.edu}}}

\begin{document}
\maketitle
\lhead{Preprint}

\begin{abstract}
\noindent 
Recent progress in frontier models has renewed interest in large-scale reinforcement learning (RL), but the cost of generating large-model rollouts makes even testing RL recipes expensive. We ask whether reasoning improvements learned by a small, weak model can benefit a larger, stronger model without using the strong model's rollouts during training. We introduce Allspark, a training and inference framework for weak-to-strong transfer through alternating chains of thought. A weak teacher is trained alongside a frozen copy of the same model; the two alternate reasoning segments, and the frozen model produces the final answer. At inference time, a stronger student replaces the frozen training partner, while both models remain fixed. Because they communicate through text, the teacher can steer students from different model families and with different tokenizers. We study Allspark at two scales: controlled Qwen experiments across math and reasoning, and larger-scale Inkling experiments on ARC-AGI-2. The Inkling experiments show accuracy gains in within-family and cross-family settings, including transfer to Kimi and Nemotron, with benefits that vary across inference settings. These findings motivate reusing a trained weak teacher across strong students and examining the resulting accuracy--token tradeoff.

\end{abstract}
\section{Introduction}
\begin{quote}
\small
\begin{CJK*}{UTF8}{gbsn}
\noindent 弟子不必不如师，师不必贤于弟子。
\end{CJK*}
\par\smallskip
\noindent\textit{A student is not necessarily inferior to the teacher,
nor is a teacher necessarily superior to the student.}
\par\smallskip
\hfill ---Han Yu, \textit{Shishuo} (``On Teachers'')
\end{quote}

Recent progress in frontier models has renewed interest in large-scale reinforcement learning (RL) \citep{team2026kimi,deepseek2026v4,xiaomi2026mimo26, microsoft2026maithinking1, openai2026gpt6astra, anthropic2026opus55}. Yet on testing an RL recipe on a large model requires repeated sampling and optimization, making even preliminary experimentation expensive. Rollout generation, particularly for long reasoning traces, can be a major bottleneck \citep{zhou2025april}. This cost limits how many training recipes researchers can test and makes it difficult for laboratories with modest compute budgets to explore improvements for frontier models. A compelling alternative is to develop reasoning capabilities in smaller models and transfer their benefits to stronger ones.

One approach is to train specialist teachers and consolidate their capabilities into a single student through on-policy distillation (OPD), as in Kimi K3 and DeepSeek-V4 \citep{team2026kimi,deepseek2026v4}. However, OPD still samples from and updates the student, retaining the expense of generating large-model rollouts when the student is large. Direct token-level distribution matching also assumes aligned tokenizations, although recent methods explicitly support OPD across different tokenizers \citep{niu2026tokenizer}. Our aim is to remove the need for the target strong student during training and transfer the learned behavior through text at inference time. This leads to the central question:

\begin{quote}
\emph{Can a weak teacher trained with RL improve a strong student without using any rollouts from that strong student during training?}
\end{quote}

We propose Allspark, a training and inference framework for weak-to-strong transfer through alternating chains of thought (Figure~\ref{fig:method-overview}). During training, two copies of a small model take turns extending a shared reasoning trace. One copy, the \emph{weak teacher}, is trained with RL; the other remains frozen and produces the final answer. At inference time, we replace the frozen training partner with a stronger student and reuse the trained teacher without further optimization. Teacher and student continue each other's reasoning, and the student produces the final answer. The transfer interface is the shared text, allowing the same teacher to steer students with different architectures and tokenizers. The teacher therefore need not be a better standalone problem solver: its role is to contribute reasoning that helps the student reach a better answer.

We study Allspark at two complementary scales. The controlled Qwen3 study uses 1.7B and 4B models \citep{yang2025qwen3} across math and reasoning to examine teacher-training choices and transfer to larger students under fixed inference limits. The larger-scale study trains an Inkling-Small teacher \citep{inklingsmall} and evaluates it on 96 development problems drawn from the public training split of ARC-AGI-2 \citep{chollet2025}. This study examines transfer within the Inkling family and across families to Kimi K2.6 \citep{moonshot2026kimi26} and Nemotron 3 Ultra \citep{nvidia2026nemotronultra}. It shows accuracy gains in various settings, providing evidence that a trained teacher can be reused across strong students. Using Inkling's thinking-effort controls \citep{tinkerthinkingeffort}, we also examine how teacher steering changes accuracy as reasoning effort varies (Figure~\ref{fig:test-time-scaling}). We report accuracy against retained completion tokens, making the scope of the token--performance comparison explicit.

Our contributions are threefold:
\begin{itemize}
    \item We introduce Allspark, which trains a weak teacher using only small-model rollouts and reuses it to steer frozen strong students through alternating chains of thought.
    \item We study teacher training and transfer at two scales: controlled Qwen experiments across two domains, and Inkling experiments with evidence of transfer within and across model families.
    \item We analyze how teacher steering interacts with test-time scaling, identifying settings in which it improves accuracy and characterizing the accompanying retained-token tradeoff.
\end{itemize}

\section{Background and Related Works}
\label{sec:background}

\paragraph{Weak supervision for a trainable strong student.}
Weak-to-strong generalization studies whether a stronger model can benefit
from a weaker supervisor without inheriting its performance ceiling.
\citet{burns2023} finetune strong models on weak labels, and \citet{yang2024}
extend this approach to selected weak-model reasoning traces. These methods
update the strong model, but supervised finetuning need not generate its
rollouts. On-policy distillation instead gives teacher feedback on sequences
sampled from the student, reducing the mismatch between training prefixes
and those encountered at inference \citep{agarwal2024gkd,lu2025opd}.
Cross-tokenizer variants can provide token-level supervision even when
teacher and student vocabularies differ \citep{niu2026tokenizer}.

Recent weak-to-strong distillation methods often transfer a change learned
by the weak model rather than its entire policy. Direct-OPD uses the
post-RL versus pre-RL log-ratio as an implicit reward on strong-student
rollouts \citep{feng2026directopd}. W2S-OPD adds a weak-model logit contrast
to the strong student's base logits and distills the resulting proxy teacher
\citep{yu2026w2sopd}. OPRD uses the weak model's policy change to rescale the
strong student's verifier-driven policy gradient \citep{park2026oprd}.
These approaches obtain strong-student rollouts and update its parameters
during transfer. They address a different resource setting from training a
weak teacher once and applying it to frozen strong students.

\paragraph{Guidance for a frozen strong student.}
Other methods keep the strong model fixed and alter how it is guided at
inference. Weak-to-strong search steers a frozen large model with likelihood
differences from tuned and untuned weak models \citep{zhou2024}. CoWeST
trains a weak collaborator using preferences derived from its interactions
with a strong model; the strong model then refines the weak model's drafts
and background information \citep{jiao2025cowest}. Co-LLM learns when to
defer to a frozen assistant \citep{shen2024}, while weak critics can prompt
a strong model to revise an answer \citep{jin2026opcd}.


\section{Allspark: Transfer Without Strong-Model Training Rollouts}
\label{sec:method}
\begin{figure}[!ht]
\centering
\includegraphics[width=\textwidth]{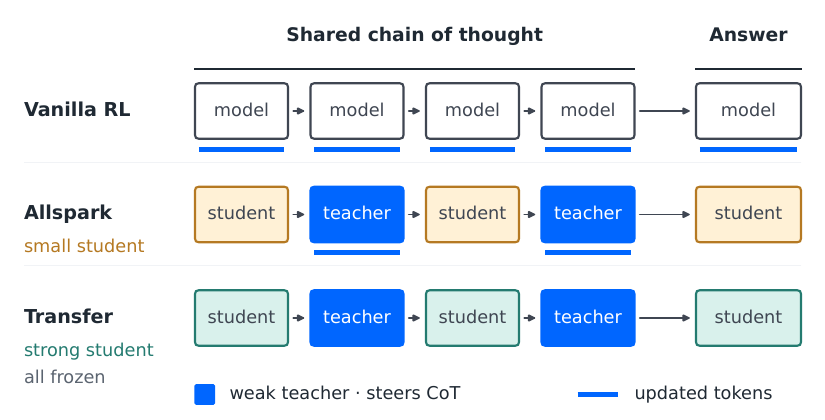}
\caption{Vanilla RL updates the full reasoning and answer. Allspark updates only the weak teacher's reasoning beside a frozen small student. Transfer freezes the teacher and replaces the student with a stronger model. Blocks show authorship; underlines mark direct token loss.}
\label{fig:method-overview}
\end{figure}

Allspark asks whether a weak model can learn to improve another model's
reasoning without using the target strong model during training. We train a
weak teacher alongside a frozen \emph{weak student}, initialized from the
same base model. They alternate reasoning chunks, and the weak student
produces the final answer. Its task reward updates only the teacher's
reasoning contributions. At inference, we freeze the teacher and replace
the weak student with a frozen strong student, which now continues the
shared reasoning and produces the answer (Figure~\ref{fig:method-overview}).
The teacher is thus trained entirely with weak-model rollouts and transferred
through text, without updating or training on rollouts from the strong
student. 

\subsection{Alternating reasoning}
Let $\pi$ be the weak teacher and $\rho$ the student. Given a problem $x$,
both models extend the same reasoning prefix $h$, initially containing $x$.
The student starts, and the two models take turns generating a chunk $z$:
\begin{equation}
z\sim
\begin{cases}
\rho(\cdot\mid h), & \text{student turn},\\
\pi(\cdot\mid h), & \text{teacher turn}.
\end{cases}
\label{eq:allspark-rollout}
\end{equation}
Each chunk is appended to $h$ before the next turn. When reasoning stops,
the student produces the final answer $y\sim \rho(\cdot\mid h)$.
Let $R(x,y)\in\{0,1\}$ indicate whether the final answer is correct.

\subsection{Training the weak teacher}
During training, $\rho$ is a frozen copy of the teacher's initial model.
Only $\pi$ is optimized, with the task objective
\begin{equation}
\max_{\pi}\ J(\pi),\qquad
J(\pi)=\mathbb E_{\pi,\rho}\bigl[R(x,y)\bigr],\qquad \rho\text{ fixed}.
\label{eq:allspark-objective}
\end{equation}
The expectation averages over training problems and the alternating rollouts
above. Only sampled teacher reasoning tokens receive direct loss; student
tokens and inserted boundary markers are masked out. The experiment-specific RL losses and training rewards are given in
Appendices~\ref{app:qwen-rl} and~\ref{app:inkling-training}.

The usefulness of a teacher chunk can be expressed as its continuation value:
\begin{equation}
Q(h,z)=\mathbb E\bigl[R(x,y)\mid h,z\bigr].
\label{eq:allspark-continuation-value}
\end{equation}
Here the teacher has emitted $z$ after prefix $h$; the expectation covers the
remaining rollout under the current $\pi$ and $\rho$. A useful chunk
therefore need not solve the problem by itself: its value comes from the
student's subsequent continuation.

\subsection{Transfer across model size and family}
At transfer, freeze the selected teacher $\pi$ and replace $\rho$ with a stronger
student. The same alternating procedure applies, with both models fixed.
Transfer changes the student's reasoning context through teacher-generated
text; it does not update the student's weights. Each writer renders and
tokenizes the shared text using its own model interface, so the models
need not share a vocabulary.

\section{Experiment}
\label{sec:experiments}
\label{sec:results}

We evaluate Allspark at two scales: controlled experiments with Qwen models and larger-scale experiments with Inkling models. The Qwen experiments examine weak-to-strong transfer through targeted ablations, while the Inkling experiments demonstrate improvements in the accuracy–token tradeoff and weak-to-strong transfer across model families.

\subsection{Controlled Qwen experiments}
\label{sec:qwen-results}
\paragraph{Setup.}
The controlled study uses two Qwen3-1.7B~\citep{yang2025qwen3} training runs: ordinary RL and Allspark, each with seed 0. Both train on the same mixed-domain pool, with equal numbers of accepted prompts from each domain per update. Allspark trains alongside a frozen Qwen3-1.7B student. The resulting 1.7B teachers are reused with frozen Qwen3-1.7B and Qwen3-4B students, without additional training. Table~\ref{tab:data} summarizes the data; Appendix~\ref{app:qwen-rl} gives the loss, training recipe, and checkpoint selection.
\begin{table}[bhtp!]
\centering\small
\caption{Qwen training and development data for the two reported domains. Both methods share the screened RL pool. Development questions are disjoint from this pool but were used for checkpoint selection and subsequent evaluation. All samples come from the datasets' original training splits.}
\label{tab:data}
\begin{tabular}{@{}llrr@{}}
\toprule
Domain & Dataset & RL pool & Development \\
\midrule
Math & MATH \citep{hendrycks2021math} & 142 & 128 \\

Reasoning & LogiQA 2.0 \citep{liu2023logiqa2} & 162 & 128 \\
\midrule
Total & & 304 & 256 \\
\bottomrule
\end{tabular}
\par\smallskip
\begin{minipage}{\textwidth}\footnotesize
Counts are distinct prompts. Each domain starts with 512 training candidates; initial-policy screening retains prompts with 1--7 correct answers in eight samples. Each update uses two accepted prompts per domain, despite the unequal pool sizes.
\end{minipage}
\end{table}

\paragraph{Controls and evaluation.}
Table~\ref{tab:main-results} compares each frozen student alone and with the checkpoint-94 Allspark teacher, an ordinary-RL teacher, and (for the 4B student) the untrained teacher. We also include the ordinary-RL-trained 1.7B policy on its own. Every paired condition uses a Qwen3-1.7B teacher; the student remains frozen and produces the final answer. We report the math and reasoning portions of the development panel in Table~\ref{tab:data}, with 128 questions per domain and one sampled answer per question. We use final-answer accuracy for both domains. Within each comparison, we match the evaluation questions, student decoding settings, and reasoning and answer limits.

\begin{table}[!ht]
\centering\small
\setlength{\tabcolsep}{8pt}
\caption{Qwen development accuracy (\%) on math and reasoning, with 128 questions per domain. The Allspark teacher is checkpoint 94; Ordinary RL is the separately trained Qwen3-1.7B policy. Dashes mark ongoing evaluations.}
\label{tab:main-results}
\begin{tabular}{@{}llrr@{}}
\toprule
Student & Teacher & Math & Reasoning \\
\midrule
Qwen3-1.7B & None & 81.3 & 73.4 \\
& Ordinary RL & 88.2 & 71.8 \\
& Allspark & 89.8 & 75.0 \\
\addlinespace
Qwen3-1.7B (RL-tuned) & None & 89.8 & 70.3 \\
\addlinespace
Qwen3-4B & None & \textbf{96.9} & 81.2 \\
& Untrained & 93.0 & 79.7 \\
& Ordinary RL & 89.8 & 79.6 \\
& Allspark & 96.1 & \textbf{82.8} \\
\bottomrule
\end{tabular}
\end{table}


\newpage
\begin{figure}[!htbp]
\centering
\includegraphics[width=0.9\textwidth]{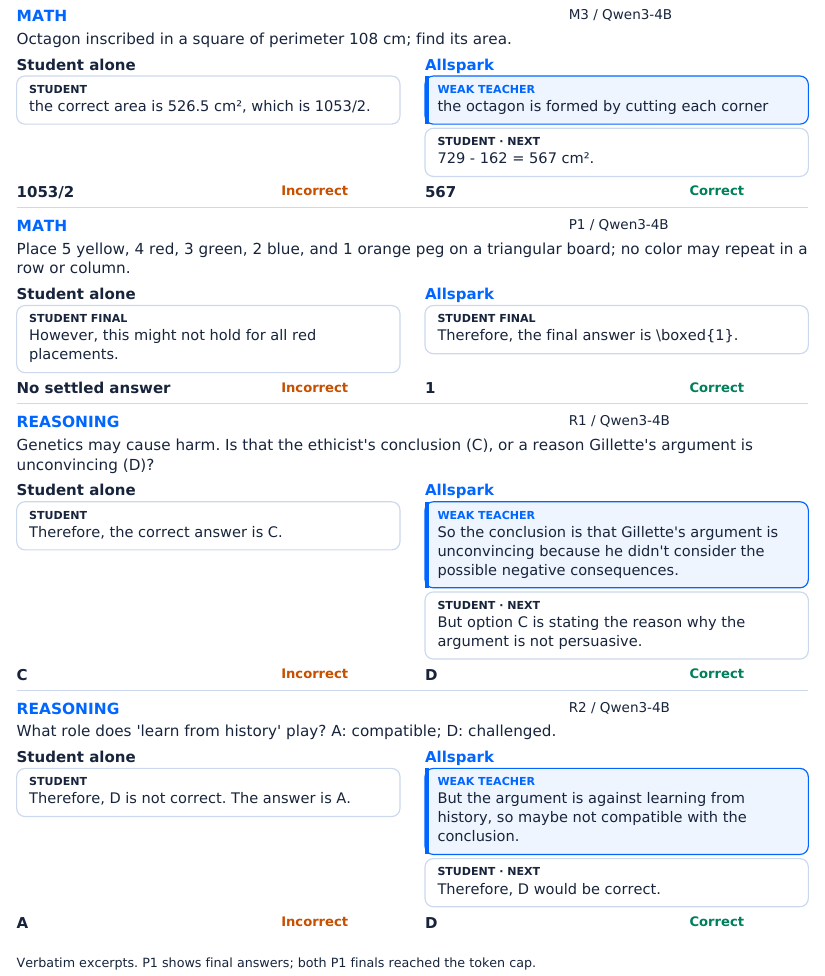}
\caption{Selected paired math and reasoning rollouts with a Qwen3-4B student and a Qwen3-1.7B weak teacher. Blue bubbles are teacher-authored; white bubbles are student-authored. M3, R1, and R2 show teacher and next-student excerpts; P1 compares final-answer excerpts. Both P1 finals hit the answer-token limit. Quotes are verbatim; task summaries are abbreviated.}
\label{fig:rollout-example}
\end{figure}

\paragraph{Takeaway.} Table~\ref{tab:main-results} provides a sign of life for weak-to-strong capability transfer from an Allspark-trained teacher across math and reasoning tasks. In the 4B ablation, using either an untrained or an ordinary-RL weak teacher to steer the stronger student lowers accuracy in both domains. Using the Allspark-trained teacher instead yields approximately neutral transfer in math and positive transfer in reasoning.

\paragraph{Case studies.}
Figure~\ref{fig:rollout-example} shows two math recoveries and two reasoning answer changes on paired Qwen3-4B development questions. In the octagon problem (M3), the student alone settles on $1053/2$; in the alternating trace, the teacher returns to corner subtraction and the student next computes the correct area, $567$. In the peg-board problem (P1), the solo final never settles on a total, whereas the alternating final states the correct answer, $1$. Both P1 finals reach the answer-token limit, so this pair identifies an outcome difference, not a single decisive handoff. In R1 and R2, the teacher reframes the relevant claim and the student's answer changes from incorrect to correct. The R2 teacher also overstates the passage. These selected traces show what an exchange can look like; Table~\ref{tab:main-results} gives the aggregate context. Appendix~\ref{app:rollout-cases} includes fuller excerpts and other cases.

\begin{figure}[!htbp]
\centering
\includegraphics[width=\textwidth]{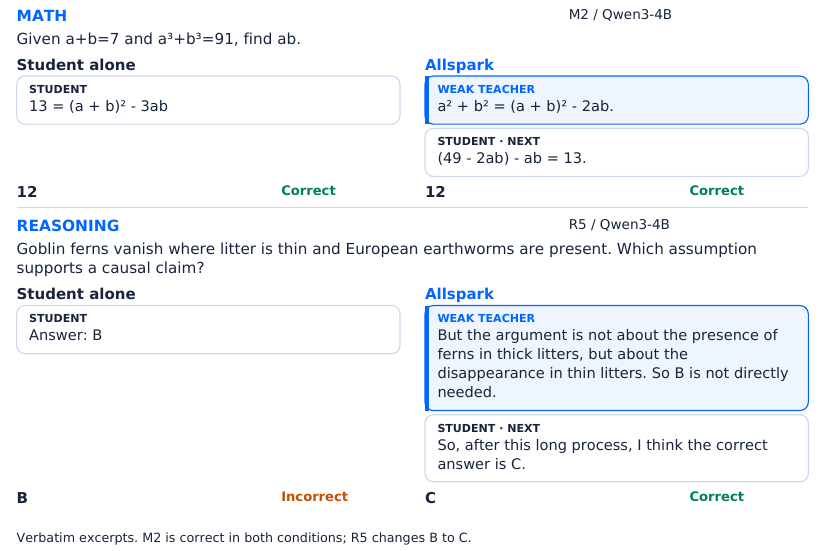}
\caption{Two further recorded handoffs with Qwen3-4B students and the same Qwen3-1.7B weak teacher. Blue bubbles are teacher-authored; other excerpts are student-authored. Both conditions solve M2; R5 changes from B (incorrect) to C (correct). Quotes are verbatim and task summaries are abbreviated.}
\label{fig:rollout-additional}
\end{figure}

Figure~\ref{fig:rollout-additional} shows two complementary behaviors. In M2, the teacher supplies an algebraic substitution that the student uses in its next turn; both conditions reach the correct answer, $12$. In R5, the student alone selects B, while the alternating rollout selects C after the teacher rules out B's stronger claim about ferns in thick litter. The student had considered C earlier, so this excerpt shows a useful continuation rather than an isolated moment of discovery.

\subsection{Inkling experiments}
\label{sec:inkling-results}
\paragraph{Setup.}
We train the weak teacher, Inkling-Small (276B total parameters, 12B active per token)~\citep{inklingsmall}, alongside a frozen copy of the same model. At evaluation, we replace the frozen partner with Inkling (975B total, 41B active)~\citep{inkling2026} as the strong student. The strong student contributes no training rollouts, and both models remain fixed during transfer. We use the 1,000 public training tasks of ARC-AGI-2~\citep{chollet2025}, split into 904 training-pool tasks and 96 development tasks. Appendix~\ref{app:inkling-training} provides the training and rollout settings.

\paragraph{Evaluation.}
Using Rl'ed teacher, we compare Inkling alone, Inkling-Small with the teacher, and Inkling with the teacher (Allspark). Each configuration is evaluated at thinking efforts $0.40$, $0.55$, $0.70$, and $0.85$, with three attempts per development task and shared task--attempt seed assignments. We report mean pass@1 over all 288 attempts and the mean retained output tokens per attempt, including both writers and the final answer.

\begin{figure}[!ht]
\centering
\includegraphics[width=\textwidth]{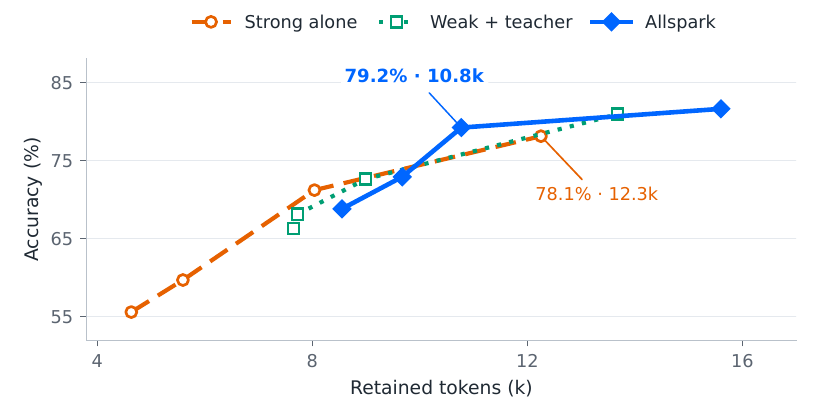}
\caption{Accuracy--token tradeoff on 96 ARC development tasks, with three attempts per task. Paired runs reuse the rl'ed teacher with an Inkling-Small or Inkling student. Points from left to right use thinking efforts 0.40, 0.55, 0.70, and 0.85. Tokens include retained output from both writers and the final answer.} 
\label{fig:test-time-scaling}
\end{figure}

\paragraph{Takeaway.}
Allspark achieves the highest observed accuracy in Figure~\ref{fig:test-time-scaling}, reaching 81.6\% compared with 78.1\% for Inkling alone. It also improves the accuracy--token tradeoff at selected operating points: Allspark reaches 79.2\% with 10.8k retained tokens, exceeding the strong student's best observed accuracy of 78.1\% at 12.3k tokens. These results support weak-to-strong transfer without updating the strong student.

\subsection{Cross-family transfer}
\label{sec:cross-family-results}
\paragraph{Setup.}
We reuse the same Inkling-Small weak teacher with two strong students from other model families: Nemotron-3-Ultra (550B total, 55B active)~\citep{nvidia2026nemotronultra}, evaluated in medium- and full-thinking modes, and Kimi-K2.6 (1T total, 32B active)~\citep{moonshot2026kimi26modelcard} on its 128K-context serving tier. Both teacher and student remain frozen; no target-family retraining is performed. Because the models use different tokenizers and reasoning formats, each handoff decodes the retained reasoning into text and re-encodes it in the next model's native reasoning channel. The student starts the trace and always produces the final answer.

Each setting compares the student alone with Allspark on the same 96 ARC development tasks, with three attempts per task and shared task--attempt seed assignments. We report mean pass@1 and task-bootstrap intervals. The teacher uses effort $0.40$ with Nemotron medium and $0.85$ with Nemotron full and Kimi; their reasoning ceilings are 24,576 and 49,152 tokens, respectively.

The alternating evaluator explicitly closes reasoning and requests a separate student answer, including after a reasoning-ceiling stop; the student-alone baseline must transition to its answer naturally within one completion. Thus, the comparison measures the complete inference procedure. Appendix~\ref{app:cross-family-protocol} specifies the text interface, stopping rules, and token accounting.

\begin{figure}[!ht]
\centering
\includegraphics[width=\textwidth]{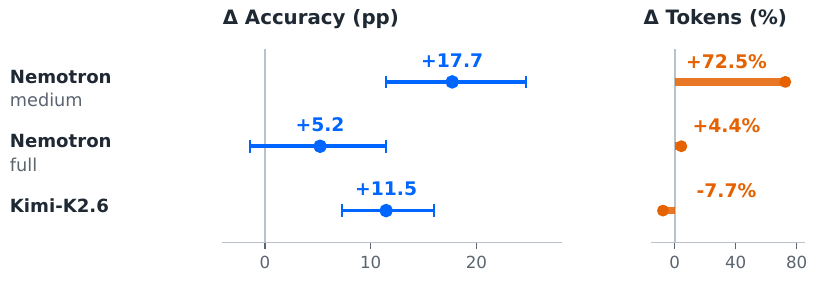}
\caption{Changes versus each student alone: Nemotron-3-Ultra (medium/full) and Kimi-K2.6 (128K). Same weak teacher; 96 ARC development tasks, three attempts each. Bars: archived 95\% task-bootstrap intervals. Tokens: retained output from both models, in native tokenizer units.}
\label{fig:cross-family-transfer}
\end{figure}

\paragraph{Transfer results.}
We evaluate the target models at native thinking settings rather than tracing an Inkling-style effort curve. Figure~\ref{fig:cross-family-transfer} shows accuracy gains with different effects on retained output length. For Kimi, Allspark improves accuracy by 11.5 percentage points while using 7.7\% fewer retained tokens. For Nemotron, mean accuracy increases by 17.7 and 5.2 percentage points in medium- and full-thinking modes, respectively, with increased token use. These results support reuse of the same teacher across model families, although accuracy and token benefits depends on the target student's inference settings.

\FloatBarrier

\section{Conclusion}

We introduced Allspark, a training and inference framework for weak-to-strong transfer through alternating chains of thought. A weak teacher learns alongside a frozen copy of the same model, then steers a stronger, frozen student at inference. This avoids strong-model rollouts during training, while the shared text interface enables transfer across model sizes and families. Our study spans math and reasoning with Qwen models and ARC-AGI-2 with larger models. The Inkling results demonstrate cross-family transfer and improved accuracy--token tradeoffs. These findings suggest a practical route to exploring RL recipes and training data with smaller models and reusing the resulting teachers with stronger students. We hope Allspark lowers the barrier to RL research and inspires extensions to model alignment and AI safety research.

\section{Limitations}
\label{sec:limitations}
Allspark avoids generating strong-model rollouts during teacher training, but inference requires serving both the teacher and the student. The additional model calls and handoffs may increase inference cost and latency, so reductions in retained tokens do not necessarily translate into lower deployment cost. Our experiments provide initial evidence under limited compute resources. Evaluation covers a restricted set of models, tasks, and training runs; broader studies are needed to establish the robustness and generality of the observed gains. Finally, our implementation relies on the ability to interrupt and resume a shared reasoning stream. Its applicability to non-thinking models, or to interfaces that do not expose a suitable reasoning stream, remains untested.



\subsection*{AI use statement}

In this work, we used generative AI tools for polishing writing, drafting experiment setting since it's mostly summarization and organization of documented experiments, and discovery of related works.
We have not used generative AI tools for research ideation, generating synthetic datasets, provide core analysis in paper and proving mathematical claims.
Additionally, we used generative AI tools for creating or modifying scientific figures or images and editing software code. We have reviewed all AI-assisted work. For example, we have gone through the citations found by AI and try our best to make sure they truly exist and they are truly relevant instead of hallucination. We take responsibility for the final content of this work,
including text, claims or artifacts produced with the aid of generative AI.

\bibliographystyle{iclr2027_conference}
\bibliography{iclr2027_conference}

\clearpage
\appendix
\section{Qwen RL implementation and training protocol}
\label{app:reproducibility}
\label{app:qwen-rl}
\subsection{Models and rollout masks}
\label{app:qwen-rl-mechanics}

Both arms initialize from \texttt{Qwen/Qwen3-1.7B} with training seed~0.
Only LoRA parameters are optimized~\citep{hu2022lora}; the pretrained weights remain fixed.
In Allspark, the student uses the same initial checkpoint without an adapter.

\paragraph{Rollouts and update masks.}
Ordinary RL samples the entire reasoning and answer from the adapted model.
Allspark starts with at most 128 student reasoning tokens, then alternates
teacher and student chunks of at most 256 tokens until reasoning ends or
reaches its shared token ceiling. Each writer conditions on the prompt and
the complete preceding stream. The frozen student generates the final answer.
The loss mask $m_t$ is:
\begin{center}
\begin{tabular}{lcc}
\hline
Token source & Ordinary RL & Allspark \\
\hline
Sampled adapted-model reasoning & 1 & 1 (teacher) \\
Sampled student reasoning & --- & 0 \\
Sampled final answer & 1 & 0 (student) \\
Forced closing token / separator & 0 & 0 \\
\hline
\end{tabular}
\end{center}
A sampled \texttt{</think>} inherits its writer's mask. A closing token
inserted at the reasoning ceiling has mask zero. Student tokens remain in
the causal context of subsequent teacher predictions. A single binary
verifier reward $r\in\{0,1\}$ for the final answer supplies the training signal;
there is no intermediate reasoning reward.

\subsection{RL loss: masked sampler-head centering}
\label{app:qwen-loss}
Both arms use the same loss. Ordinary RL applies it to the adapted model's
reasoning and answer tokens; Allspark applies it only to the teacher's
reasoning tokens. The calculation has three steps.

\paragraph{Step 1: solution advantage.}
For one prompt, sample $G$ trajectories and grade their final answers
with binary rewards $r_1,\ldots,r_G$. Compute the group's mean $\mu$ and
population standard deviation $\sigma$, then assign each trajectory its
standardized reward:
\begin{equation}
\begin{aligned}
\mu &= \frac{1}{G}\sum_{j=1}^{G}r_j,\qquad
\sigma^2 = \frac{1}{G}\sum_{j=1}^{G}(r_j-\mu)^2,\\
A_i &=
\begin{cases}
(r_i-\mu)/\sigma,&\sigma>0,\\
0,&\sigma=0.
\end{cases}
\end{aligned}
\label{eq:qwen-group-advantage}
\end{equation}
We use $G=32$ trajectories per prompt. The same $A_i$ is used at every
trainable position in trajectory $i$.
A better-than-group-average answer has positive advantage; a worse answer
has negative advantage. No epsilon is added to the standard deviation.

\paragraph{Step 2: loss at one trainable token.}
Let $h$ be the full prefix and $y$ the sampled next token. The current
trainable model assigns probability $p_\theta(v\mid h)$ to token $v$.
The stored distribution $q(v\mid h)$ comes from the checkpoint that sampled
$y$, evaluated at this same prefix $h$. It is held fixed and may come from
an older checkpoint. Let $H$ contain its $K$ highest-probability tokens,
and let $A$ be this trajectory's advantage from Step~1. For this position, minimize
\begin{equation}
\ell(\theta)=-A\left[
\underbrace{\mathbf{1}\{y\in H\}\log p_\theta(y\mid h)}_{\text{sampled-token score}}
-\underbrace{\sum_{v\in H}q(v\mid h)\log p_\theta(v\mid h)}_{\text{sampler-weighted center}}
\right].
\label{eq:qwen-token-loss}
\end{equation}
The first term scores the token that was sampled. The second is a weighted
sum of the learner's log probabilities over the saved sampler head. For an
in-head sample, positive advantage favors the sampled-token score relative
to this center; negative advantage reverses that direction. If $y\notin H$,
the first term is zero and only the centering term remains.

Only $p_\theta$ receives gradients. The advantage, sampler probabilities,
and membership in $H$ are fixed. The probabilities in $H$ retain their
original mass: they are not renormalized to sum to one. The head is used
to compute the loss; generation itself does not use top-$K$ truncation.
We use $K=128$. This is a head-only variant of score centering~\citep{marek2026scorecentering}. Unlike the published implementation, which retains the sampled-token log probability outside the head and models the tail with the learner distribution rescaled to the sampler's tail mass, our loss sets the sampled-token term to zero when $y\notin H$ and omits the tail.

\paragraph{Step 3: loss for one optimizer update.}
An update contains $M$ prompt groups of $G$ trajectories each.
Let $\mathcal B$ be this batch of $MG$ trajectories and
$\mathcal T(\tau)$ the trainable positions
in trajectory $\tau$. The update loss is
\begin{equation}
\mathcal L(\theta)=\frac{1}{MG}
\sum_{\tau\in\mathcal B}\ \sum_{t\in\mathcal T(\tau)}
\ell_{\tau,t}(\theta).
\label{eq:qwen-sampler-head-loss}
\end{equation}
We use $M=6$ prompt groups (two per domain), giving 192 trajectories per
update. Thus token losses are \emph{summed}, and trajectories are \emph{averaged}.
There is no division by a trajectory's token count. The set $\mathcal T$
implements the mask in Appendix~\ref{app:qwen-rl-mechanics}: student and
forced tokens are excluded from the Allspark loss. A trajectory with no
trainable positions contributes zero loss but still enters its group's
advantage calculation and the denominator $MG$.

\subsection{Asynchronous RL and filtering}
\label{app:qwen-async}
Three rollout actors generate trajectories while one learner updates the
adapter. Each actor holds one immutable adapter snapshot for a prompt group,
including screening and the fresh training rollouts. The learner pools six
accepted groups (two per domain), accumulates their gradients, performs one
optimizer step, and publishes a new adapter checkpoint. A batch can therefore
contain groups sampled by different behavior-policy versions.

Immediately before consumption, each group's version lag must satisfy
$0\le v_{\mathrm{learner}}-v_{\mathrm{sampler}}\le\Delta_{\max}$, where
$\Delta_{\max}$ is the maximum permitted lag, set to two updates. Older
queued groups are discarded. Each loss-bearing token uses its own saved sampler probabilities
in Equation~\ref{eq:qwen-token-loss}. This is asynchronous off-policy
optimization with bounded lag, without an importance-sampling ratio, PPO
clipping, reference-policy KL term, or entropy bonus. The sampler-head term
is not an importance-weight correction for stale rollouts.

\paragraph{Prompt and group filtering.}
Both arms use the same frozen training pool drawn from MATH and
LogiQA2. Initial-policy solo screening retains a prompt when its first
complete eight-sample group has 1--7 successes. Online, each actor pins a
sampling snapshot, screens a candidate with eight fresh trajectories, and
requires 1--7 successes. It then draws 32 fresh trajectories using a separate
random-seed namespace; groups with 1--31 successes are retained in full.
Screening trajectories and tied training groups do not enter updates.
Each batch contains six distinct prompts, two per domain; prompts may recur
in later updates. Allspark shares student random-seed keys within a group,
including the answer stage, while teacher keys vary across trajectories.

\subsection{Solution-length controls and termination}
\label{app:qwen-length}
Length is controlled by separate hard generation ceilings for reasoning and
the final answer. The reasoning ceiling is shared across all chunks from
both writers, not allocated afresh to each writer or turn. Each call is
limited by the remaining budget; Allspark additionally limits its first
student chunk to 128 tokens and later chunks to 256. All returned raw tokens
are retained in the shared stream, without paragraph trimming or prefix
cropping.

\begin{center}
\begin{tabular}{lrr}
\toprule
Limit (tokens) & Training / historical evaluation & Revised evaluation \\
\midrule
Total context & 8,192 & 16,384 \\
Shared reasoning & 4,096 & 8,192 \\
Final answer & 1,536 & 4,096 \\
\bottomrule
\end{tabular}
\end{center}

A sampled \texttt{</think>} ends reasoning naturally. If the reasoning
ceiling is reached first, the generator inserts \texttt{</think>} with zero
loss mask and starts the answer phase. An EOS or empty generation during
reasoning instead invalidates the answer stage. The answer stops at its own
ceiling or native EOS; unused reasoning budget is not transferred to the
answer allowance. Before generation, the full prompt, both allowances, and
required boundary tokens must fit the context window; an oversized request
is rejected rather than silently cropped.

There is no minimum generation length, length-dependent reward penalty, or
per-sequence length normalization in the loss. Equation~\ref{eq:qwen-sampler-head-loss}
sums trainable-token terms, so different trajectory lengths contribute
different numbers of terms. A cap hit is recorded, but does not itself force
zero reward: the saved final answer is scored by the verifier. Report actual
reasoning, answer, and trainable-token counts and both cap rates; identical
ceilings do not imply identical realized length or compute. The longer
evaluation protocol changes inference only and cannot repair truncated
training rewards.

\subsection{Optimization settings}
Gradients accumulate one trajectory at a time over the accepted batch,
followed by one clipped optimizer step. Log probabilities and loss arithmetic
use FP32; the base model is loaded in BF16. Causal forward passes retain
all prefixes preceding loss-bearing tokens, with gradient checkpointing.
Vocabulary projections are evaluated in chunks of 32 loss-bearing positions.
\begin{table}[!ht]
\centering\small
\caption{Shared settings for the local Qwen training runs.}
\label{tab:qwen-recipe}
\begin{tabular}{ll}
\hline
Setting & Value \\
\hline
Optimizer & AdamW; constant learning rate $10^{-5}$ \\
Adam moments / epsilon & $(0.9,0.999)$ / $10^{-8}$ \\
Weight decay / gradient-norm ceiling & $0$ / $1$ \\
LoRA rank $r$ / $\alpha$ / dropout & $16$ / $32$ / $0$ \\
LoRA projections & $q,k,v,o$, gate, up, down; no biases \\
Training context ceiling & 8,192 tokens \\
Reasoning / answer ceilings & 4,096 / 1,536 generated tokens \\
Sampling & Temperature $1$; top-$p=1$; top-$k$ disabled \\
Sampler head used in the loss & 128 tokens; no generation truncation \\
Queue capacity & 12 prompt groups, four per domain \\
\hline
\end{tabular}
\end{table}

\FloatBarrier
\subsection{Development evaluation and checkpoint selection}
\label{app:qwen-selection}
The frozen development panel contains 384 questions, with 128 each from the
MATH, TACO, and LogiQA~2.0 training splits, disjoint from the local training
pool. Each candidate is evaluated as a teacher alternating with the same
frozen, unadapted Qwen3-1.7B student; the student writes the final answer.
There is one sampled answer per question (seed 0), using the revised
evaluation limits in Appendix~\ref{app:qwen-length} and the same sampling
settings and chunk schedule. The training limits remain unchanged.

\paragraph{Checkpoint selection rule.}
All 384 development
questions are evaluated for each candidate. Let $N_d$ be the number of
development questions in domain $d$ and $c_d(s)$ the number correct at
checkpoint $s$. Selection is
\begin{equation}
 s^*=\min\underset{s\in\mathcal C}{\operatorname{argmax}}
 \frac{c_{\mathrm{math}}(s)+c_{\mathrm{reasoning}}(s)}{N_{\mathrm{math}}+N_{\mathrm{reasoning}}}.
 \label{eq:qwen-selection}
\end{equation}
Here $N_{\mathrm{math}}=N_{\mathrm{reasoning}}=128$. Exact ties choose the earlier
update. Reasoning accuracy and the
three-domain mean are reported separately and do not choose the checkpoint.
Ordinary-RL update 92 is the nominated comparison checkpoint.
.
\begin{table}[!ht]
\centering\small
\caption{Finalized Allspark checkpoints under the revised evaluation protocol. Entries are correct answers on the two reported domains; each has 128 development questions. Checkpoint 94 is the fixed selection from the original three-domain protocol. Bold marks the selected checkpoint. These are development results, not public-test scores.}
\label{tab:qwen-selection}
\begin{tabular}{@{}lrr@{}}
\toprule
Checkpoint & Math & Reasoning \\
\midrule
55 & 116 & 93 \\
91 & 114 & 96 \\
\textbf{94} & 119 & 96 \\
\bottomrule
\end{tabular}
\end{table}
The completed review selects checkpoint 94 (Table~\ref{tab:qwen-selection});
it is held fixed for subsequent evaluations and transfer to the 4B
student. Ordinary-RL checkpoint 92 retains its earlier nomination.

\subsection{Development learning curves}
\label{app:qwen-curves}
\begin{figure}[!ht]
\centering
\includegraphics[width=\textwidth]{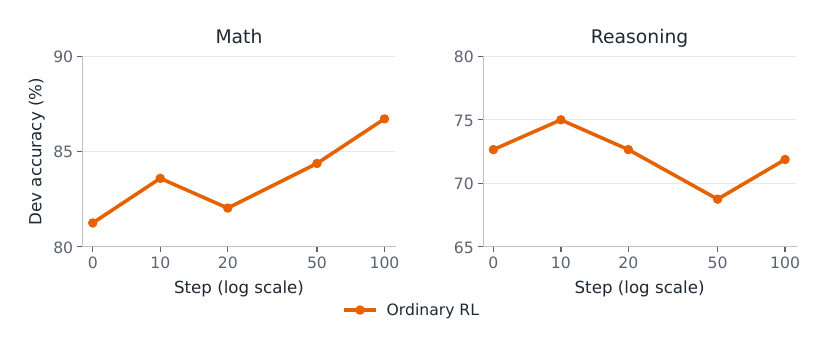}
\caption{Development accuracy of the ordinary-RL-trained Qwen3-1.7B teacher
alternating with the frozen student, under the historical 8K protocol
(4,096 reasoning and 1,536 answer tokens). Each domain has 128 fixed questions;
the figure reports math and reasoning only. Filled circles connect the initial checkpoint
(step 0) to checkpoints 10, 20, 50, and 100 without smoothing.}
\label{fig:qwen-dev-curves}
\end{figure}

\clearpage
\subsection{Training diagnostics}
\label{app:qwen-training-curves}
\begin{figure}[!ht]
\centering
\includegraphics[width=\textwidth]{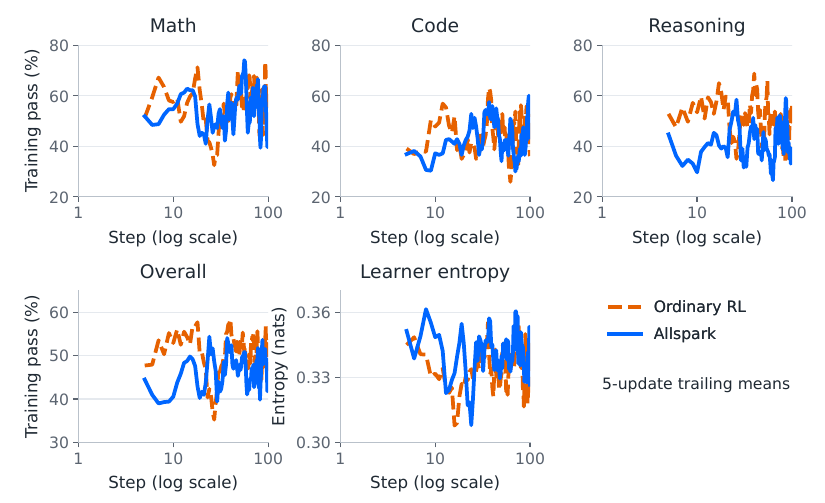}
\caption{Qwen3-1.7B training diagnostics over the first 100 committed updates
of each method on a logarithmic step axis (one seed per method). Lines show full trailing-five-update means, starting at update 5. Pass rates are measured on
accepted, filtered training rollouts; overall is the equal-domain mean.
Learner entropy is measured on the respective trainable-token contexts.
The accepted prompts and rollout policies differ between methods, so these
curves are descriptive training diagnostics, not held-out accuracy or paired
treatment effects.}
\label{fig:qwen-learning-curves}
\end{figure}
Allspark completed 100 updates; the ordinary-RL run stopped at 110.
Figure~\ref{fig:qwen-learning-curves} shows the common 1--100 range. The
additional ten ordinary-RL updates remain part of its training expenditure.
Update counts do not equate generated tokens, optimized tokens, or GPU time;
screening and rejected rollouts also contribute to training cost.

\FloatBarrier

\clearpage
\section{Inkling training protocol}
\label{app:tinker}

\subsection{Inkling training and inference}
\label{app:inkling-training}
\paragraph{Training.}
We trained a rank-$32$ LoRA adapter~\citep{hu2022lora} on Inkling-Small for $72$ rollout steps
(indexed $0$--$71$), paired with a frozen copy of the same base model.
The frozen copy opened each reasoning trace and generated the visible answer.
Each step sampled $16$ tasks at all three thinking efforts,
$\{0.4,0.55,0.7\}$, with $32$ rollouts per task--effort group: nominally
$1{,}536$ training rollouts. We used four minibatches, keeping each group
together, with one optimizer update per nonempty minibatch. The learning rate
was $2\times10^{-5}$ through step $24$ and $5\times10^{-5}$ thereafter.

\paragraph{RL loss.}
Because training used the Tinker API, we chose its built-in
\texttt{importance\_sampling} policy-gradient objective~\citep{tinkerimportance}.
Each trajectory contributes
\begin{equation}
\ell(\pi)=-\frac{\bar n}{n}\,A\sum_{t\in T}
\frac{\pi(y_t\mid h_t)}{\pi_{\mathrm{old}}(y_t\mid h_t)},
\qquad A=r-\bar r .
\label{eq:inkling-rl-loss}
\end{equation}
Here $\pi$ is the current teacher, $\pi_{\mathrm{old}}$ is the teacher that
sampled the rollout, and $h_t$ is the full shared prefix before token $y_t$.
$T$ contains only teacher-generated reasoning positions, $n=|T|$, and
$\bar n$ is the mean teacher-token count across retained trajectories in the
rollout batch. The factor $\bar n/n$ normalizes each trajectory's contribution
by its trainable length. The baseline $\bar r$ is the mean reward within the
same task--effort group; we do not divide by the group standard deviation.
We sum these losses within each minibatch. The prompt, frozen-model chunks,
and final answer receive no policy-gradient loss.

\paragraph{Reward and length control.}
Let $s$ indicate exact task success, $c$ denote cell accuracy, and $L$ count
retained reasoning tokens from both writers, excluding the final answer.
We use
\begin{equation}
r=w s+(1-w)c-
\min\!\left\{p_{\max},\lambda\max\!\left(0,\frac{L}{L_0}-1\right)\right\}.
\label{eq:inkling-length-reward}
\end{equation}
We set $w=0.85$, $L_0=5{,}000$, and $p_{\max}=0.30$.
Steps $0$--$35$ use $\lambda=0$; steps $36$--$71$ continue the same adapter
with $\lambda=0.15$. Thus the length penalty is zero through $5{,}000$
reasoning tokens, then increases by $0.15$ per additional $5{,}000$ tokens,
up to $0.30$. Training also imposes a hard reasoning ceiling of $24{,}576$
tokens; on reaching it, the reasoning channel is closed and the frozen
model generates the final answer.

\paragraph{Models and rollout settings.}
All Inkling models are served through Tinker. The teacher and its frozen training partner use \path{thinkingmachines/Inkling-Small}; the strong transfer student is \path{thinkingmachines/Inkling}. A deterministic task-ID hash defines the 904/96 training-pool/development split. The student opens each trace with at most 256 tokens, followed by alternating teacher and student chunks of at most 2,048 tokens. Handoffs are trimmed to the last paragraph boundary when at least half the sampled chunk can be retained. The frozen student produces the final answer as a JSON object containing the predicted output grids. Both writers use temperature $1.0$ and top-$p$ $1.0$, with a final-answer allowance of 4,096 tokens. In the checkpoint-63 effort sweep, the reasoning ceiling is 24,576 tokens at efforts $0.40$, $0.55$, and $0.70$, and 49,152 tokens at effort $0.85$.

\subsection{Cross-family inference protocol}
\label{app:cross-family-protocol}
\paragraph{Models and operating points.}
The cross-family evaluation pairs the Inkling-Small teacher from the within-family study with frozen Nemotron-3-Ultra or Kimi-K2.6 students. Neither the teacher nor the target student is updated during these evaluations. Table~\ref{tab:cross-family-settings} lists the three operating points; the two Nemotron rows are settings of one model family. The numerical effort parameter applies only to Inkling-Small. The target students instead use their native thinking renderers, so equal numerical effort or equal reasoning compute across families is not assumed.

\begin{table}[!ht]
\centering\small
\caption{Cross-family inference settings. Context and reasoning limits are in tokens. The context limit applies to the student; the reasoning guard is checked in the receiving model's tokenizer at each handoff.}
\label{tab:cross-family-settings}
\begin{tabular}{lrrr}
\toprule
Student setting & Teacher effort & Student context & Reasoning limit \\
\midrule
Nemotron-3-Ultra, medium & 0.40 & 65,536 & 24,576 \\
Nemotron-3-Ultra, full & 0.85 & 65,536 & 49,152 \\
Kimi-K2.6, 128K & 0.85 & 131,072 & 49,152 \\
\bottomrule
\end{tabular}
\end{table}

\paragraph{Text handoffs.}
Each model receives the same ARC task messages, rendered in its own format. Every reasoning chunk is decoded with its writer's tokenizer, stripped of model-specific control markers, and appended to a shared text trace. Before the next turn, the complete trace is re-encoded with the receiving model's tokenizer and placed in its native hidden-reasoning channel. This preserves the retained reasoning text while flattening model-specific message boundaries; token IDs are never passed directly between families. The student starts the trace, subsequent turns alternate between teacher and student, and the student always generates the final answer. The evaluation configuration specifies a 256-token first chunk, 2,048-token later chunks, paragraph trimming that retains at least half of a sampled chunk, temperature $1.0$, top-$p$ $1.0$, and a final-answer allowance of 4,096 tokens.

\paragraph{Stopping and grading.}
At each handoff, the evaluator recomputes the trace length in the receiving model's tokenizer and limits the next chunk by both the reasoning ceiling and the remaining context, reserving space for the answer. Either model can end reasoning with its native closing marker. Alternation also stops when the reasoning allowance is exhausted. The evaluator then closes the hidden channel in the student's format and requests a separate final answer, including after a reasoning-ceiling stop. By contrast, the student-alone baseline generates one uninterrupted completion with a total allowance equal to its reasoning room plus the answer allowance; it must emit its own hidden-to-visible boundary. Only the visible final answer is graded against the exact ARC output grids.

\paragraph{Aggregation and scope.}
Each condition contains 288 attempts over the same 96 development tasks. Mean pass@1 averages exact success over attempts; it is not best-of-three accuracy. For student-alone comparisons, task--attempt seed assignments are paired. The archived 95\% intervals use 5,000 paired bootstrap resamples of task-level mean differences, with bootstrap seed 17. The Inkling-Small--teacher reference is reused from the within-family scan; its medium-effort comparison is task matched but not seed matched. Token totals sum retained teacher and student output, including the final answer, in each writer's native tokenizer units; discarded chunk suffixes are excluded. These totals are neither a common-tokenizer budget nor total billed generation. A separate 16-task target-model pilot examined thinking modes and context limits, including an earlier 32K Kimi configuration.

\subsection{Earlier within-family control}
\label{app:inkling-controls}
This evaluation uses checkpoint 35 at effort $0.7$, with one attempt on each of the 96 development tasks. Each frozen student (Inkling-Small or Inkling) is evaluated alone, with an untrained Inkling-Small teacher, and with the checkpoint-35 Allspark teacher. All model weights are fixed during evaluation.

\begin{table}[ht]
\centering\small
\caption{Step 35, effort 0.7, one attempt on each of 96 ARC development tasks.}
\label{tab:historical}
\begin{tabular}{llr}
\toprule
Student & Teacher & Accuracy (\%) \\
\midrule
Inkling-Small & None & 60.4 \\
Inkling-Small & Untrained & 60.4 \\
Inkling-Small & Allspark-trained & 70.8 \\
Inkling & None & 75.0 \\
Inkling & Untrained & 66.7 \\
Inkling & Allspark-trained & 76.0 \\
\bottomrule
\end{tabular}
\end{table}

\subsection{Dollar cost and latency}
\label{app:tinker-cost-latency}
\paragraph{Cost model.}
We report the inference-cost estimates preserved in the Tinker experiment archive. They use the rates in Table~\ref{tab:tinker-rate-assumptions}, which apply the archive's recorded 50\% account discount to its list-price snapshot. These are historical analysis assumptions, not current price quotes. The calculation assumes a separate prefix cache for each writer: each model prefills the prompt on its first turn, pays the cached-read rate for its existing prefix on later turns, and pays fresh-prefill rates for newly received text. Retained output, including the final answer, is charged at the generating model's sampling rate.

\begin{table}[!ht]
\centering\small
\caption{Historical rate assumptions, in USD per million tokens.}
\label{tab:tinker-rate-assumptions}
\begin{tabular}{lrrr}
\toprule
Model / serving tier & Fresh prefill & Cached read & Sampling \\
\midrule
Inkling-Small & 0.580 & 0.058 & 1.440 \\
Inkling & 1.870 & 0.187 & 4.680 \\
Nemotron-3-Ultra, 64K & 2.490 & 0.249 & 6.225 \\
Kimi-K2.6, 128K & 2.575 & 0.515 & 6.405 \\
\bottomrule
\end{tabular}
\end{table}

\begin{table}[!ht]
\centering\small
\caption{Archived modeled inference cost, in USD per attempt, over 96 development tasks with three attempts each. Weak + teacher always uses an Inkling-Small student and the trained teacher; the cross-family rows reuse that reference. For the Inkling rows, effort applies to both writers; for cross-family rows it denotes the teacher's effort, with the target student using its native thinking setting.}
\label{tab:tinker-inference-cost}
\begin{tabular}{lrrrr}
\toprule
Student setting & Effort & Student alone & Weak + teacher & Allspark \\
\midrule
Inkling & 0.40 & 0.0266 & 0.0198 & 0.0483 \\
 & 0.55 & 0.0311 & 0.0199 & 0.0546 \\
 & 0.70 & 0.0426 & 0.0232 & 0.0606 \\
 & 0.85 & 0.0623 & 0.0354 & 0.0895 \\
\midrule
Nemotron, medium & 0.40 & 0.0459 & 0.0196 & 0.0757 \\
Nemotron, full & 0.85 & 0.1230 & 0.0352 & 0.1407 \\
Kimi, 128K & 0.85 & 0.1466 & 0.0352 & 0.1774 \\
\bottomrule
\end{tabular}
\end{table}

At the same Inkling effort, Allspark's modeled cost per attempt is $1.42$--$1.82\times$ that of the student alone. Across different efforts, the highlighted Inkling comparison uses 12.1\% fewer retained tokens (10,771 versus 12,253), but only about 2.7\% less modeled cost (\$0.0606 versus \$0.0623). The cross-family estimates are higher for Allspark in all three settings. In particular, Kimi's 7.7\% reduction in retained tokens accompanies a 21.1\% increase in modeled cost, from approximately \$0.1466 to \$0.1774 per attempt. Prefix processing, cached rereads, and the mix of teacher and student tokens therefore affect the dollar comparison.

\clearpage
\section{Rollout case studies}
\label{app:rollout-cases}
\noindent The cases below are selected paired development rollouts. They illustrate the content exchanged between a frozen student and the weak teacher; they do not establish the causal effect of an individual teacher span. We include successful, unchanged, ambiguous, and unsuccessful pairs.

\paragraph{Selection context.} The cases include the two domains reported in Table~\ref{tab:main-results} and two supplementary coding rollouts. They cover unchanged, recovered, and unsuccessful outcomes. Full task identities, excerpt role offsets, and source hashes are retained in the accompanying case manifest; complete traces remain in the run archive.

\begin{center}\small
\begin{tabular}{llll}\toprule Case & Domain & Student & Native $\rightarrow$ Allspark \\ \midrule
M1 & Math & 4B & correct $\rightarrow$ correct \\
M2 & Math & 4B & correct $\rightarrow$ correct \\
M3 & Math & 4B & incorrect $\rightarrow$ correct \\
R1 & Reasoning & 4B & incorrect $\rightarrow$ correct \\
R2 & Reasoning & 4B & incorrect $\rightarrow$ correct \\
C1 & Coding & 4B & incorrect $\rightarrow$ correct \\
C4 & Coding & 4B & correct $\rightarrow$ incorrect \\
\bottomrule\end{tabular}
\end{center}

\clearpage
\begin{center}
\begin{minipage}{\textwidth}
\captionsetup{hypcap=false}
\centering
\includegraphics[page=1,width=\textwidth]{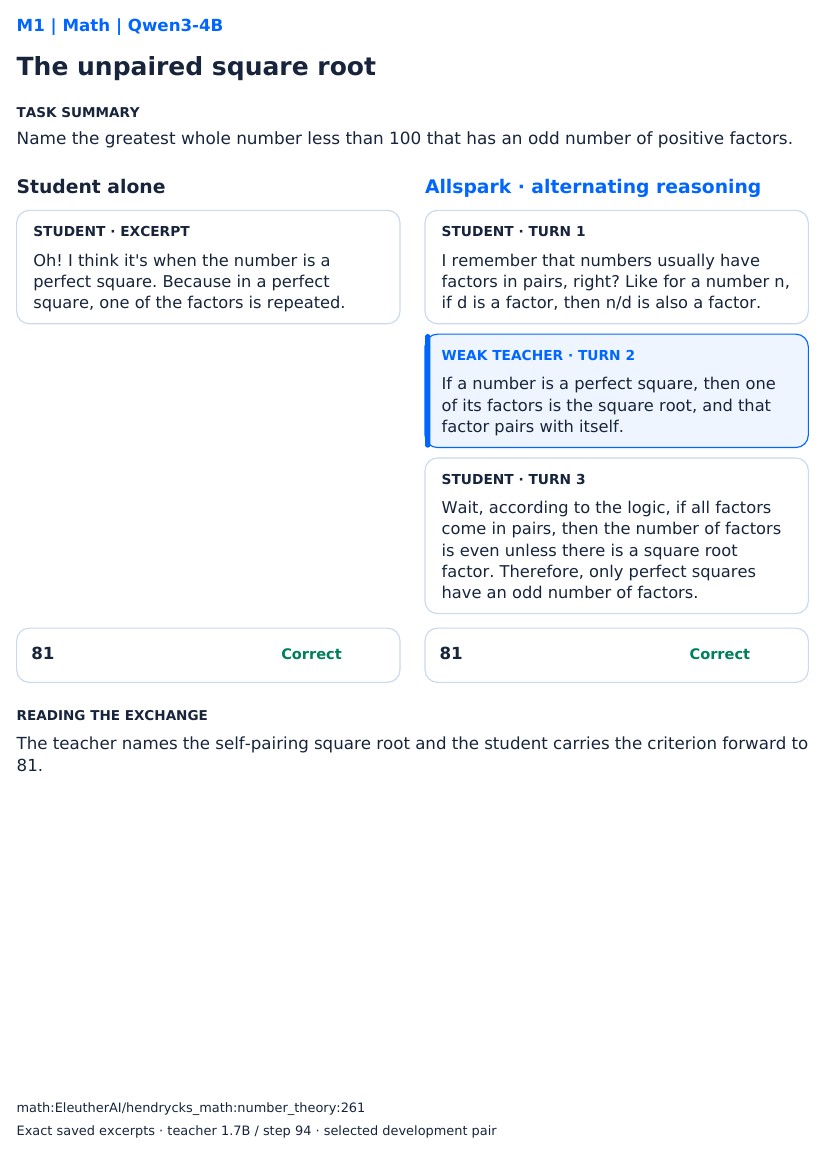}
\captionof{figure}{Case M1: The unpaired square root. Recorded excerpts; task text is summarized above.}
\label{fig:case-m1}
\end{minipage}
\end{center}

\clearpage
\begin{center}
\begin{minipage}{\textwidth}
\captionsetup{hypcap=false}
\centering
\includegraphics[page=2,width=\textwidth]{tex/allspark_case_studies.pdf}
\captionof{figure}{Case M2: Eliminate the square terms. Recorded excerpts; task text is summarized above.}
\label{fig:case-m2}
\end{minipage}
\end{center}

\clearpage
\begin{center}
\begin{minipage}{\textwidth}
\captionsetup{hypcap=false}
\centering
\includegraphics[page=3,width=\textwidth]{tex/allspark_case_studies.pdf}
\captionof{figure}{Case M3: Recovery after uncertain geometry. Recorded excerpts; task text is summarized above.}
\label{fig:case-m3}
\end{minipage}
\end{center}

\clearpage
\begin{center}
\begin{minipage}{\textwidth}
\captionsetup{hypcap=false}
\centering
\includegraphics[page=9,width=\textwidth]{tex/allspark_case_studies.pdf}
\captionof{figure}{Case R1: Separate a reason from the conclusion. Recorded excerpts; task text is summarized above.}
\label{fig:case-r1}
\end{minipage}
\end{center}

\clearpage
\begin{center}
\begin{minipage}{\textwidth}
\captionsetup{hypcap=false}
\centering
\includegraphics[page=10,width=\textwidth]{tex/allspark_case_studies.pdf}
\captionof{figure}{Case R2: Find the claim being challenged. Recorded excerpts; task text is summarized above.}
\label{fig:case-r2}
\end{minipage}
\end{center}

\clearpage
\begin{center}
\begin{minipage}{\textwidth}
\captionsetup{hypcap=false}
\centering
\includegraphics[page=5,width=\textwidth]{tex/allspark_case_studies.pdf}
\captionof{figure}{Case C1: Parity before an exception. Recorded excerpts; task text is summarized above.}
\label{fig:case-c1}
\end{minipage}
\end{center}

\clearpage
\begin{center}
\begin{minipage}{\textwidth}
\captionsetup{hypcap=false}
\centering
\includegraphics[page=8,width=\textwidth]{tex/allspark_case_studies.pdf}
\captionof{figure}{Case C4: A shared indexing error. Recorded excerpts; task text is summarized above.}
\label{fig:case-c4}
\end{minipage}
\end{center}

\end{document}